%% file: main.tex
\documentclass[11pt]{article}
\PassOptionsToPackage{hyperfootnotes=false}{hyperref}

\usepackage[final]{acl}

\usepackage{times}
\usepackage{latexsym}
\usepackage[T1]{fontenc}
\usepackage[utf8]{inputenc}
\usepackage{microtype}
\usepackage{inconsolata}
\usepackage{graphicx}
\usepackage{booktabs}
\usepackage{amsmath}
\usepackage{amssymb}
\usepackage[table]{xcolor}

\usepackage[most]{tcolorbox}
\usepackage{listings}

\lstdefinestyle{promptstyle}{
  basicstyle=\ttfamily\footnotesize,
  breaklines=true,
  breakatwhitespace=false,
  columns=fullflexible,
  keepspaces=true,
  showstringspaces=false,
  tabsize=2
}

\newtcblisting{promptbox}{
  colback=gray!4,
  colframe=gray!45,
  arc=1mm,
  boxrule=0.4pt,
  left=1mm,
  right=1mm,
  top=1mm,
  bottom=1mm,
  listing only,
  listing options={style=promptstyle}
}

\title{Calibrating Overconfidence Without Sacrificing Confidence: Probe-Conditioned Head Intervention for LLMs}

\author{
Ke Li\textsuperscript{1,2}\thanks{These authors contributed equally.}
\quad
Chongzhe Zhang\textsuperscript{1,3}\footnotemark[1]
\quad
Zifan Zeng\textsuperscript{1,4}
\quad
Feng Liu\textsuperscript{1}
\quad
Qunli Zhang\textsuperscript{1}
\quad
Zheng Hu\textsuperscript{1}
\\
\textsuperscript{1}Huawei Heisenberg Research Center
\quad
\textsuperscript{2}EPFL
\quad
\textsuperscript{3}TU Berlin
\quad
\textsuperscript{4}TUM
}

\begin{document}
\maketitle

\begin{abstract}
Large language models often express high confidence in answers that are wrong. Standard calibration remedies typically act globally or at the score level, reducing unwarranted confidence but also risking erosion of warranted confidence on correct answers. We introduce \emph{Probe-Conditioned Head Intervention} (PCHI), an inference-time method that uses a frozen probe to detect likely wrong-but-confident responses and conditionally rescales downstream attention-head outputs during confidence generation. On Qwen3-4B-Instruct solving OpenMathInstruct problems with a structured binary confidence field, readout-token PCHI converts 82.2\% of originally wrong-yes confidence readouts to \texttt{no}, while a joint intervention across upstream confidence-template tokens reduces ECE from 21.9\% to 9.2\% and damages only 5.1\% of originally correct-yes readouts. The readout-token effect also appears on Gemma3-4B, though upstream interventions are weaker and more mask-dependent. These results show that verbalized overconfidence can be selectively reduced through conditionally applied internal intervention, partially decoupling the suppression of unwarranted confidence from the loss of warranted confidence.
\end{abstract}


\input{latex/introduction}
 

\input{latex/related}


\input{latex/method}


\input{latex/experiments}


\input{latex/results}

\input{latex/conclusion}

\bibliography{custom}

\appendix

\input{latex/appendix}

\end{document}

%% file: latex/introduction.tex
\section{Introduction}
\label{sec:intro}

For a language model to be useful in decision-making, it is not enough that its
answers are often correct; a system or person acting on those answers must also
be able to tell when to trust them \citep{guo2017calibrationmodernneuralnetworks}.
This makes the reliability of a model's expressed confidence an important
concern. Yet large language models often express high confidence even when their
answers are wrong, making verbalized confidence a poor guide to correctness
\citep{xiong2024llmsexpressuncertaintyempirical, geng-etal-2024-survey}.

A natural response is to recalibrate. Standard post-hoc methods such as
temperature scaling \citep{guo2017calibrationmodernneuralnetworks} apply a
single adjustment to every response, and instruction- or RLHF-tuned models may
inherit reward signals that favor high confidence regardless of correctness
\citep{leng2025tamingoverconfidencellmsreward}. Such methods can improve
aggregate calibration, but they do not explicitly target the
wrong-but-confident subset. Pushing confidence down far enough to suppress
wrong-but-confident responses can also reduce warranted confidence on correct
answers, a tension documented both for classical networks
\citep{joy2022sampledependentadaptivetemperaturescaling} and for language
models \citep{xie2024calibratinglanguagemodelsadaptive}.

This motivates a more selective form of calibration: one that acts only when a
response is likely to be wrong-but-confident. Conditional and adaptive
interventions already provide evidence that model behavior can be controlled
non-uniformly, for example in refusal
\citep{lee2025programmingrefusalconditionalactivation}, cross-lingual transfer
\citep{maraia-etal-2026-activation}, and two-stage confidence steering
\citep{miao2026closingconfidencefaithfulnessgaplarge}. However, these methods
do not directly isolate wrong-but-confident responses through an intervention on
the internal computation that produces the final confidence readout.

Recent mechanistic work suggests that such an internal intervention is
plausible. Verbalized confidence appears to be formed around answer-adjacent or
template positions and read out later
\citep{kumaran2026llmscomputeverbalconfidence}, while inflated confidence has
been linked to specific late-position components
\citep{zhao2026wiredoverconfidencemechanisticperspective}. If evidence about whether a
confident answer is actually correct is represented before the final confidence
value is emitted, then an intervention conditioned on that evidence may reduce
unwarranted confidence without globally suppressing all confidence.

We instantiate this idea as \emph{probe-conditioned head intervention} (PCHI).
A frozen probe reads a hidden state on the confidence template and estimates
whether the response is likely to be wrong-but-confident. The probe score then
gates a learned rescaling of downstream attention-head outputs, so the
intervention is applied in proportion to wrong-confident evidence rather than
uniformly across all responses.

%% file: latex/related.tex
\section{Related Work}
\label{sec:related}

\paragraph{Calibrating language model confidence.}
A long line of work seeks to align a model's confidence with its accuracy. One
line elicits confidence directly from the model in natural language, and finds
that such verbalized confidence can be a meaningful but imperfect signal of
correctness \citep{lin2022teachingmodelsexpressuncertainty}; relatedly, models retain some ability to self-evaluate whether their own
answers are correct \citep{kadavath2022languagemodelsmostlyknow}. A second line adjusts confidence post hoc. Methods such as temperature scaling
\citep{guo2017calibrationmodernneuralnetworks} fit a single parameter on a
held-out set and rescale all predictions uniformly; in language models this
miscalibration is exacerbated by RLHF, whose reward models favor confident
responses \citep{leng2025tamingoverconfidencellmsreward}. Because a uniform
rescaling moves every prediction in the same direction, it cannot simultaneously
raise confidence where it is too low and lower it where it is too high -- a
limitation noted for classical networks
\citep{joy2022sampledependentadaptivetemperaturescaling} and addressed in
language models by predicting an input-dependent temperature
\citep{xie2024calibratinglanguagemodelsadaptive}. These methods operate on the
output confidence distribution. Closer to our setting, recent work probes
internal representations to \emph{estimate} confidence, for example from the
stability of perturbed representations \cite{khanmohammadi-etal-2025-calibrating}. Our intervention differs in kind: rather than estimating a confidence score, it
acts inside the model to \emph{modify} the confidence readout, and is applied
only to the responses a probe judges wrong-but-confident, so that warranted
confidence on the remaining responses is left intact.

\paragraph{Activation interventions and conditional control.}
A growing body of work steers model behavior by modifying internal
representations, typically by adding a fixed direction to the residual stream
\citep{turner2024steeringlanguagemodelsactivation, li2024inferencetimeinterventionelicitingtruthful, zou2025representationengineeringtopdownapproach}. Such
unconditional steering is known to lack selectivity, affecting related behaviors
together \citep{wehner2025taxonomyopportunitieschallengesrepresentation}. Conditional variants make the intervention
input-dependent: conditional activation steering gates whether to steer in order
to program refusal \citep{lee2025programmingrefusalconditionalactivation}, and
probe-gated steering selects a class-specific transformation after a probe
predicts the class \citep{maraia-etal-2026-activation}. Causal head gating, in
contrast, learns a fixed, input-independent scalar gate per attention head from a
next-token objective, in order to assign heads a causal role for interpretability
\citep{nam2025causalheadgatingframework}. Our method shares the mechanism of
per-head gating, but differs in two ways: the gate is modulated per response by a
frozen probe's continuous estimate of wrong-confidence, and it is deployed to
selectively correct verbalized overconfidence rather than to characterize head
roles.

\paragraph{The mechanism of verbalized confidence.}
Recent work asks where and how verbalized confidence is computed. Confidence
appears to be formed at answer-adjacent positions and read out later
\citep{kumaran2026llmscomputeverbalconfidence}; inflated confidence has been
attributed to specific components writing at late token positions
\citep{zhao2026wiredoverconfidencemechanisticperspective}; and the direction
governing calibration has been found nearly orthogonal to the one governing
verbalized confidence \citep{miao2026closingconfidencefaithfulnessgaplarge}.
A complementary line shows that a model's internal states carry decodable
evidence of whether its own answer is correct, recoverable by simple probes \citep{azaria2023internalstatellmknows, macdiarmid2024sleeperagentprobes}. These analyses are largely descriptive, and where they intervene, they do so
to characterize the mechanism rather than to selectively correct errors. They
nonetheless motivate our approach: if correctness-relevant information is present
internally and partly separable from the expressed-confidence signal, then a
conditional internal intervention should be able to act on the former without
uniformly disturbing the latter.

%% file: latex/method.tex
\section{Method}
\label{sec:method}

\begin{figure*}[t]
\centering
\includegraphics[width=\textwidth]{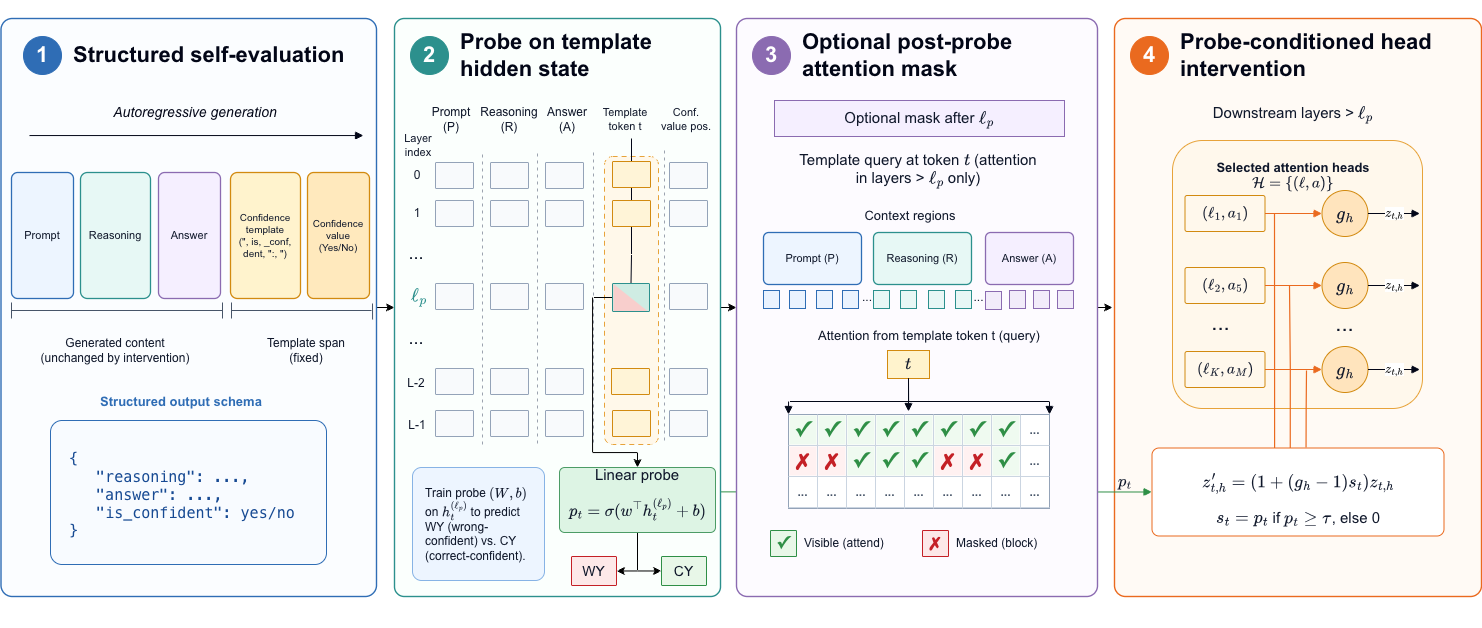}
\caption{Overview of probe-conditioned head intervention (PCHI). \textbf{(1)} The model produces a structured self-evaluation output. \textbf{(2)} At a confidence-template token~$t$, a frozen linear probe reads the hidden state at layer~$\ell_p$ and predicts whether the response is wrong-but-confident (WY) rather than correct-confident (CY), yielding a score $p_t$. \textbf{(3)} An optional attention mask, applied only in layers after $\ell_p$, restricts which context regions (prompt, reasoning, answer) the template query at~$t$ may attend to; layers up to and including $\ell_p$ retain the original attention, so the probe score is unaffected. \textbf{(4)} In downstream layers, each attention head is rescaled by a learned coefficient $g_h$, most of which are kept near $1$ by a sparsity penalty, with strength conditioned on the probe score: $z'_{t,h}=(1+(g_h-1)s_t)\,z_{t,h}$ where $s_t=p_t$ if $p_t\ge\tau$ and $0$ otherwise. The intervention thus fires only on responses the probe flags as wrong-confident.}
\label{fig:pipeline}
\end{figure*}

Probe-conditioned head intervention corrects verbalized overconfidence at the moment the model prepares to emit its confidence value. The method has three conceptual components: a structured self-evaluation format that exposes a binary confidence readout, a confidence-template probe that detects latent evidence of wrong-confident responses, and a probe-conditioned head-level intervention that is applied only when this evidence is present. This design treats overconfidence correction as a selective control problem: the model should revise confidence on wrong answers while leaving warranted confidence largely unchanged.

The central design principle is to separate \emph{detection} from \emph{intervention}. The probe estimates whether a confident response is likely to be wrong, while the learned head parameters specify how downstream attention heads should be rescaled when such evidence is detected. Separating these roles gives the method a direct selectivity target: convert wrong-confident responses from \texttt{yes} to \texttt{no} without globally suppressing correct-confident responses.

We intervene only after the model has produced its reasoning and answer. Intervening during the reasoning or answer span can alter the hidden-state trajectory that supports problem solving, potentially moving the model into regions where its subsequent reasoning and final answer are unreliable. In contrast, the confidence template is a fixed span between the answer and the confidence value. At inference time, these template tokens can be forced rather than sampled, so the intervention changes the computation that determines the final self-evaluation while leaving the generated answer and the template token sequence fixed.

\subsection{Structured Self-Evaluation and Readout}
\label{sec:method-setting}

We study verbalized confidence under a fixed JSON output schema. For each input problem $x_i$, the model is instructed to generate a response with three fields (Appendix ~\ref{sec:appendix-prompt-template}):
\[
\{\texttt{reasoning},\ \texttt{answer},\ \texttt{is\_confident}\}.
\]
The field \texttt{answer} contains the final answer, and \texttt{is\_confident} contains a binary self-evaluation, either \texttt{yes} or \texttt{no}. Let $a_i$ denote the parsed answer and let $c_i\in\{0,1\}$ indicate whether $a_i$ is correct under the task evaluator. Let $r_i\in\{\texttt{yes},\texttt{no}\}$ be the parsed confidence value. Each response therefore belongs to one of four groups: correct-yes (CY), correct-no (CN), wrong-yes (WY), and wrong-no (WN).

Our intervention targets WY responses. A WY response is problematic because the model gives an incorrect answer while explicitly reporting high confidence. A useful intervention should therefore reduce confidence on WY examples without broadly suppressing confidence on CY examples. We measure the confidence readout with the yes-no logit gap
\begin{equation}
\label{eq:confidence-gap}
\Delta_i
=
\operatorname{LSE}_{v\in V_{\texttt{yes}}} o_{i,v}
-
\operatorname{LSE}_{v\in V_{\texttt{no}}} o_{i,v},
\end{equation}
where $\operatorname{LSE}$ denotes log-sum-exp, $o_i$ is the next-token logit vector at the confidence-value prediction position, and $V_{\texttt{yes}}$ and $V_{\texttt{no}}$ are the candidate tokens for the two confidence values. Positive $\Delta_i$ favors \texttt{yes}; negative $\Delta_i$ favors \texttt{no}.

\subsection{Confidence-Template Evidence Probes}
\label{sec:method-probes}

The confidence template is the token span after the \texttt{answer} value and before the \texttt{is\_confident} value. Let $\mathcal{T}_i$ denote the aligned template coordinates for example $i$. For a template coordinate $t\in\mathcal{T}_i$ and transformer layer $\ell$, let $h_{i,t}^{(\ell)}\in\mathbb{R}^{d}$ be the hidden state at that coordinate after layer $\ell$.

We train position- and layer-specific linear probes to detect wrong-confident evidence. For each template coordinate $t$ and layer $\ell$, the probe predicts whether a confident response is wrong:
\begin{equation}
\label{eq:probe-score}
p_{i,t}^{(\ell)}
=
\sigma\!\left((w_{t,\ell})^\top h_{i,t}^{(\ell)} + b_{t,\ell}\right),
\end{equation}
where positive examples are WY responses and negative examples are CY responses. We use calibrated linear probes so that the resulting score can be interpreted as a graded estimate of wrong-confident evidence while keeping the detector simple enough to run online.

The probe layer $\ell_p$ must satisfy two requirements: the probe should separate WY from CY reliably, and there must be downstream layers available for intervention in the same forward pass. We restrict the intervention to layers after $\ell_p$ so that the hidden states consumed by the probe remain on the same distribution as those used during probe training. This is important because the probe begins to separate WY from CY in middle layers: modifying earlier layers would change the probe input itself and could make the probe score unreliable. The resulting intervention therefore uses the evidence detected at $\ell_p$ to modulate only later head outputs at the same template coordinate.

\subsection{Probe-Conditioned Head Intervention}
\label{sec:method-intervention}

Probe-conditioned head intervention rescales attention-head outputs only when the probe indicates likely wrong-confident evidence. Let $\mathcal{H}$ be the set of intervention heads, where each head $h=(\ell,a)$ consists of layer $\ell$ and attention-head index $a$, and all $\ell$ are downstream of $\ell_p$. For each $h\in\mathcal{H}$, we learn a scalar intervention parameter $g_h$, initialized at the identity value $1$. Unlike a global ablation, this parameter is not applied uniformly to every response; its strength is controlled by the probe score at the current confidence-template coordinate.

Let $z_{i,t,h}$ be the pre-output-projection attention output of head $h$ at template coordinate $t$. We replace this head output with
\begin{equation}
\label{eq:head-intervention}
z'_{i,t,h}
=
\left(1 + (g_h - 1)s_{i,t}\right) z_{i,t,h},
\end{equation}
where $s_{i,t}\in[0,1]$ is the probe-conditioned intervention strength. If $s_{i,t}=0$, then $z'_{i,t,h}=z_{i,t,h}$ and the model is unchanged. If $s_{i,t}=1$, then the head is fully scaled by its learned coefficient $g_h$. Intermediate values interpolate between the original computation and the computation with learned scaling. This formulation allows the coefficient to suppress or amplify a head, depending on the learned value of $g_h$, while preserving exact identity behavior when the probe does not activate.

During intervention-parameter training, we use soft probe conditioning, $s_{i,t}=p_{i,t}^{(\ell_p)}$. Soft conditioning makes the learning objective differentiable with respect to the learned coefficients and uses the probe score as a graded measure of wrong-confident evidence. During inference, we use a hard selection rule with threshold $\tau=0.5$:
\begin{equation}
\label{eq:runtime-gate}
s_{i,t}
=
\begin{cases}
p_{i,t}^{(\ell_p)}, & p_{i,t}^{(\ell_p)}\ge \tau,\\
0, & p_{i,t}^{(\ell_p)}<\tau.
\end{cases}
\end{equation}
Thus, the probe makes a binary decision about whether to intervene, but the magnitude of the intervention above threshold remains proportional to the probe probability.
This inference rule uses only the probe score and does not require correctness labels.

The method can be applied at one or more confidence-template coordinates. Let $\mathcal{S}\subseteq\mathcal{T}_i$ be the set of intervention coordinates. In the single-coordinate setting, the method learns intervention parameters for one selected coordinate at a time. In the multi-coordinate setting, the same rule is applied independently as generation passes through each coordinate in $\mathcal{S}$.

\subsection{Learning Selective Interventions}
\label{sec:method-learning}

The base language model and the confidence-template probes are frozen while learning the scalar intervention parameters. Intervention learning uses examples from the WY and CY groups. For a training example $i$, let $\Delta_i$ be the original yes-no logit gap and let $\Delta'_i$ be the gap after applying probe-conditioned head intervention. Since positive gaps favor \texttt{yes}, the objective should push WY examples below a target margin while keeping CY examples close to their original confident state.

We optimize the following hinge-style objective:
\begin{equation}
\label{eq:training-objective}
\begin{aligned}
\mathcal{L}
&=
\mathcal{L}_{\mathrm{WY}}
+
\mathcal{L}_{\mathrm{CY}}
+
\lambda\sum_{h\in\mathcal{H}} |g_h-1|,\\
\mathcal{L}_{\mathrm{WY}}
&=
\mathbb{E}_{i\in \mathrm{WY}}
\left[\max(0,\Delta'_i - m)\right],\\
\mathcal{L}_{\mathrm{CY}}
&=
\mathbb{E}_{i\in \mathrm{CY}}
\left[\max(0,\rho\Delta_i - \Delta'_i)\right].
\end{aligned}
\end{equation}
The first term penalizes WY examples whose post-intervention gap remains above the target $m$. The second term protects CY examples by requiring the post-intervention gap to remain at least a fraction $\rho$ of the original gap. The final term is an identity regularizer: head parameters that are unnecessary for the selective correction objective are encouraged to stay near $1$, limiting unnecessary changes to the model and encouraging a sparse set of influential heads without treating sparsity itself as a head-selection claim.

Intervention learning is performed on fixed replays of the confidence-template span, which lets us compare $\Delta_i$ and $\Delta'_i$ at the same confidence-value prediction point. At runtime, the probe score at layer $\ell_p$ controls only downstream heads at the same coordinate in the current forward pass.

The intervention can also include an optional post-probe attention mask. We partition the pre-template context into the problem prompt $\mathcal{P}_i$, generated reasoning field $\mathcal{R}_i$, and generated answer field $\mathcal{A}_i$, and choose a visible context $C_i$ such as $\mathcal{P}_i$, $\mathcal{R}_i$, $\mathcal{A}_i$, or $\mathcal{R}_i\cup\mathcal{A}_i$. Full context corresponds to no restriction. When the mask is enabled, layers up to and including $\ell_p$ keep the original attention pattern, while template queries in layers $\ell>\ell_p$ attend only to $C_i$ among the pre-template tokens. Thus the probe input distribution is unchanged, and the learned head intervention is applied only after the probe score has been produced.

%% file: latex/experiments.tex
\section{Experiments}
\label{sec:setup}

\paragraph{Experimental Setup.}
We evaluate on OpenMathInstruct\citep{toshniwal2024openmath2}, following prior preprocessing\citep{nam2025causalheadgatingframework} by filtering invalid examples before selecting 5,000 problems from the original training split and 5,000 problems from the original validation split. For each problem, the model produces one structured JSON response with greedy decoding (temperature 0). We use the training split to collect hidden states on the six-token confidence template and train calibrated diagonal-LDA probes to distinguish wrong-yes from correct-yes examples.

Table~\ref{tab:group-counts} reports the baseline group composition on the validation split for each model. Both models exhibit a high \texttt{yes} confidence rate, so the evaluation is dominated by the correct-yes and wrong-yes groups targeted by our intervention.

\begin{table}[t]
\centering
\small
\setlength{\tabcolsep}{4pt}
\begin{tabular}{lrrrrrr}
\toprule
Model & CY & WY & CN & WN & Acc. & Yes \\
\midrule
Qwen3-4B  & 3808 & 1067 & 28 &  97 & 76.7 & 97.5 \\
Gemma3-4B & 3527 & 1298 & 19 & 156 & 70.9 & 96.5 \\
\bottomrule
\end{tabular}
\caption{
Baseline validation-set group composition. CY, WY, CN, and WN denote correct-yes, wrong-yes, correct-no, and wrong-no responses, respectively.
}
\label{tab:group-counts}
\end{table}
We evaluate two instruction-tuned models: Qwen3-4B-Instruct and Gemma3-4b-it. We select the probe layer using training-set AUROC heatmaps over confidence-template positions and transformer layers (Appendix~\ref{sec:appendix-probe-diagnostics}). For Qwen, layer 18 is used as the probe layer and layers 19--22 as intervention layers, giving 128 learned head coefficients. For Gemma, layer 16 is used as the probe layer and layers 17--20 as intervention layers, giving 32 learned head coefficients. These probe layers lie in the middle part of each model, where WY/CY separability is already strong (check ~\ref{sec:appendix-probe-diagnostics} for detailed explanation), while leaving downstream layers available for intervention. This choice is also consistent with prior findings that confidence-relevant answer-evaluative information is represented in middle-to-late transformer layers \citep{kumaran2026llmscomputeverbalconfidence}. Intervention parameters are trained only on wrong-yes and correct-yes examples. We train Qwen and Gemma both with batch size 8, learning rate 0.04, step 200, L1 weight 0.05, hinge ratio 0.7, and random seed 42.

\paragraph{Evaluation Metrics.}
We evaluate whether an intervention reduces unwarranted confidence while preserving warranted confidence. Let $g_i\in\{\mathrm{CY},\mathrm{CN},\mathrm{WY},\mathrm{WN}\}$ be the baseline group of example $i$, and let $\Delta'_i$ be the post-intervention yes-no logit gap at the confidence-value position. We classify the post-intervention confidence as \texttt{yes} when $\Delta'_i>0$ and \texttt{no} otherwise. The wrong-yes correction rate measures how often originally wrong-yes confidence readouts are converted to \texttt{no}:
\[
\mathrm{WY\ Corr.}
=
\frac{
\sum_i \mathbf{1}[g_i=\mathrm{WY}]\,\mathbf{1}[\Delta'_i\le 0]
}{
\sum_i \mathbf{1}[g_i=\mathrm{WY}]
}.
\]
The correct-yes damage rate measures how often originally correct-yes confidence readouts are converted to \texttt{no}:
\[
\mathrm{CY\ Dmg.}
=
\frac{
\sum_i \mathbf{1}[g_i=\mathrm{CY}]\,\mathbf{1}[\Delta'_i\le 0]
}{
\sum_i \mathbf{1}[g_i=\mathrm{CY}]
}.
\]
We report calibration with expected calibration error (ECE). We define the model's verbal confidence as the yes/no-restricted probability of \texttt{yes} at the confidence-value position. Let $q_i$ denote this probability:
\[
s_i^{\texttt{yes}}=\operatorname{LSE}_{v\in V_{\texttt{yes}}}o_{i,v},
\qquad
s_i^{\texttt{no}}=\operatorname{LSE}_{v\in V_{\texttt{no}}}o_{i,v},
\]
\[
q_i
=
\frac{\exp s_i^{\texttt{yes}}}
{\exp s_i^{\texttt{yes}}+\exp s_i^{\texttt{no}}}.
\]
Given $M=10$ equal-width confidence bins $\{B_m\}_{m=1}^{M}$ over $[0,1]$ and
correctness labels $y_i\in\{0,1\}$, ECE is
\[
\mathrm{ECE}
=
\sum_{m=1}^{M}
\frac{|B_m|}{n}
\left|
\frac{1}{|B_m|}\sum_{i\in B_m}y_i
-
\frac{1}{|B_m|}\sum_{i\in B_m}q_i
\right|.
\]
Finally, we report AUROC using answer correctness as the positive label and $q_i$ as the score. Because Qwen3-4B reports \texttt{yes} on 97.5\% of validation examples, the wrong-yes and correct-yes groups account for nearly all of the data, so this AUROC primarily reflects discrimination within the subset that the intervention targets.

%% file: latex/results.tex
\section{Results}
\label{sec:results}

\subsection{Selective Calibration Is Achievable at the Readout}
\label{sec:existence}

We first evaluate the most direct intervention point, the final confidence-template token whose forward pass produces the yes/no confidence logits. As shown in Table~\ref{tab:readout}, PCHI substantially improves both calibration and discrimination at this readout position. On Qwen3-4B, ECE drops from 21.9 to 11.7 and AUROC rises from 66.5 to 90.3, while 82.2\% of wrong-yes readouts are converted to \texttt{no}. On Gemma3-4B, ECE drops from 26.3 to 16.9 and AUROC rises from 76.8 to 81.7, with 52.5\% of wrong-yes readouts converted. This selectivity comes with measurable correct-yes damage, 10.8\% on Qwen3-4B and 11.6\% on Gemma3-4B.

The activation-steering baseline uses the same probe layer as PCHI, but replaces learned head coefficients with a single hidden-state shift. For the selected confidence-template token, we compute a normalized mean-difference direction on the training set from wrong-yes activations to wrong-no activations at that layer, and add a scaled version of this direction to the current-token hidden state during inference. We sweep the scaling coefficient over $[1, 8]$ and report the best result. At the readout token, this baseline leaves ECE and AUROC close to the no-intervention baseline and converts almost no wrong-yes readouts. This contrast suggests that the improvement is not explained by a generic mean-difference shift at the same layer.

\input{tables/table1}

\begin{figure}[t]
\centering
\includegraphics[width=\columnwidth]{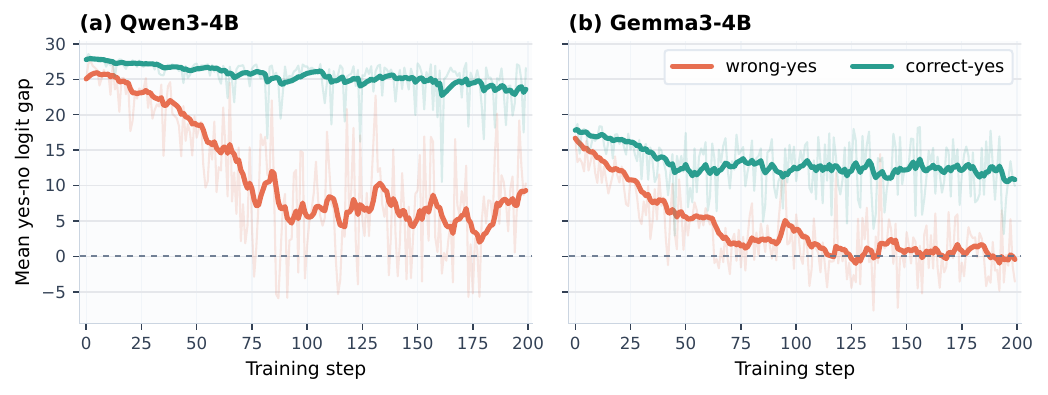}
\caption{Training dynamics of the yes-no logit gap for the readout-position PCHI runs in Table~\ref{tab:readout}. Each panel shows the mean post-intervention gap on wrong-yes and correct-yes training examples. Faint curves are raw per-step means, and bold curves are exponential moving averages. The dashed horizontal line marks the yes/no decision boundary.}
\label{fig:training-gap}
\end{figure}

The training dynamics provide a direct diagnostic of this selectivity. Figure~\ref{fig:training-gap} plots the mean post-intervention yes-no logit gap during training for the same readout-position PCHI runs as Table~\ref{tab:readout}. In both models, the wrong-yes gap decreases sharply over training, while the correct-yes gap changes more modestly and remains on the confident side of the yes/no boundary. This behavior matches the intended objective: the learned intervention mainly reduces the margin supporting unwarranted confidence, rather than uniformly suppressing all responses that initially read out \texttt{yes}.

\subsection{Selectivity Along the Answer-to-Readout Pathway}
\label{sec:reach}
This experiment tests whether PCHI is limited to the readout token or can also act at upstream confidence-template tokens. This distinction matters because upstream confidence-template tokens are generated after the answer but before the final yes/no confidence readout, so successful intervention there would show that selective calibration is not confined to the last confidence-token computation. We therefore apply the same probe-conditioned head intervention at individual upstream confidence-template tokens and at a joint set of upstream tokens, while leaving the generated answer unchanged.

Table~\ref{tab:reach} reports PCHI at individual upstream confidence-template tokens and at the joint token set 1--5. For Qwen3-4B, tokens 1 and 2 have little effect, token 3 worsens discrimination, and tokens 4 and 5 become substantially stronger. The joint intervention over tokens 1--5 gives the best overall balance, reducing ECE to 9.2 and raising AUROC to 91.1 while correcting 63.3\% of wrong-confident examples with 5.1\% correct-yes damage. For Gemma3-4B, single upstream-token interventions are much weaker: tokens 1--4 correct less than 1\% of wrong-confident examples, and token 5 corrects only 4.5\%. The joint tokens 1--5 intervention is more useful than any single upstream token, improving AUROC from 76.8 to 82.1 and lowering ECE from 26.3 to 23.6, but it still corrects only 9.5\% of wrong-confident examples.

\input{tables/table2}

The weak upstream-token results under full context suggest possible attenuation along the template-to-readout path. Unlike the readout token, an upstream-token intervention must affect later computations through attention, and its contribution can be reduced when subsequent template positions attend over the full context. To test this possibility without changing the probe input, we apply the attention mask only after the probe layer and restrict post-probe template queries to a specified visible context. Table~\ref{tab:mask} evaluates this control on Qwen3-4B for tokens 1--3. With prompt-only visibility, tokens 1 and 2 become strong intervention points, reaching 70.2\% and 80.6\% wrong-yes correction with large AUROC gains. In contrast, token 3 remains ineffective even under the prompt-only mask, correcting only 0.9\% of wrong-confident examples. This outcome aligns with the probe diagnostics in Appendix~\ref{sec:appendix-probe-diagnostics}, where token 3 shows weaker WY/CY separability than neighboring template positions.

\input{tables/table3}

\begin{figure}[t]
\centering
\includegraphics[width=\columnwidth]{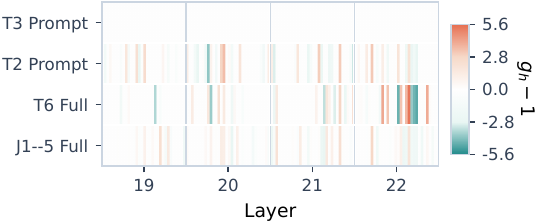}
\caption{Learned head coefficients for representative Qwen3-4B intervention settings. Rows are settings and columns are attention heads grouped by layers 19--22; each cell shows $g_h-1$, with white indicating identity behavior. T3 Prompt and T2 Prompt apply PCHI under prompt-only post-probe visibility at tokens 3 and 2, respectively. T6 Full is the readout-token setting, and J1--5 Full denotes joint intervention over tokens 1--5. Active-head counts with $|g_h-1|>0.5$ are 0, 32, 35, and 48, respectively.}
\label{fig:qwen-head-coefficients}
\end{figure}

The learned coefficients provide a complementary diagnostic for why the same post-probe visibility restriction helps some template positions but not others. Figure~\ref{fig:qwen-head-coefficients} visualizes representative Qwen3-4B settings. Token 3 with prompt-only visibility remains essentially at the identity, whereas token 2 under the same visibility learns sparse non-identity coefficients. The readout token and the joint tokens 1--5 setting also learn sparse non-identity patterns. This contrast supports the interpretation that attention masking does not by itself create correction; it can expose an effective intervention path only when the selected template coordinate provides usable wrong-confident evidence.

Gemma3-4B shows that context restriction can substantially strengthen joint upstream-token correction while preserving most correct-yes cases. Table~\ref{tab:gemma-mask} evaluates the joint tokens 1--5 intervention while varying the visible context. The full-context setting is conservative, with only 9.5\% wrong-yes correction and 0.7\% correct-yes damage. Restricting visibility to the answer field yields the strongest correction and best ECE, raising wrong-yes correction to 40.4 and lowering ECE to 17.3. This setting lowers AUROC relative to full context and increases correct-yes damage to 6.1\%, but the damage increase is limited compared with the 30.9-point gain in wrong-yes correction, and AUROC remains above the no-intervention baseline in Table~\ref{tab:readout}.

\input{tables/table4}

\subsection{Ablation Study}
\label{sec:ablations}

Finally, we test whether the attention mask alone explains the upstream-token gains. Table~\ref{tab:mask-only} compares a prompt-only mask without head intervention to PCHI under the same visible-context restriction on Qwen3-4B. Masking alone slightly worsens ECE and AUROC relative to no intervention and corrects only 1.0\% of wrong-confident examples. Under the same prompt-only context, PCHI at token 1 corrects 70.2\% of wrong-confident examples and PCHI at token 2 corrects 80.6\%. Token 3, however, remains close to the mask-only behavior. These controls indicate that the gains require both the learned head intervention and a template coordinate with usable wrong-confident evidence. The attention mask by itself is insufficient.

\input{tables/table5}

%% file: tables/table1.tex
\begin{table}[t]
  \centering
  \footnotesize
  \setlength{\tabcolsep}{3pt}
  \resizebox{\columnwidth}{!}{%
  \begin{tabular}{@{}lrrrr@{}}
    \toprule
    Method & ECE $\downarrow$ & AUROC $\uparrow$ & WY Corr. $\uparrow$ & CY Dmg. $\downarrow$ \\
    \midrule
    \rowcolor{black!5}
    \multicolumn{5}{l}{\textit{Qwen3-4B}} \\
    No intervention & 21.9 & 66.5 & --- & --- \\
    Activation steering & 21.9 & 66.7 & 0.1 & 0.0 \\
    \textbf{PCHI (ours)} & \textbf{11.7} & \textbf{90.3} & \textbf{82.2} & 10.8 \\
    \addlinespace
    \rowcolor{black!5}
    \multicolumn{5}{l}{\textit{Gemma3-4B}} \\
    No intervention & 26.3 & 76.8 & --- & --- \\
    Activation steering & 26.2 & 76.8 & 0.5 & 0.0 \\
    \textbf{PCHI (ours)} & \textbf{16.9} & \textbf{81.7} & \textbf{52.5} & 11.6 \\
    \bottomrule
  \end{tabular}
  }
  \caption{Readout-token intervention results. All values are percentages.}
  \label{tab:readout}
\end{table}

%% file: tables/table2.tex
\begin{table}[t]
  \centering
  \footnotesize
  \setlength{\tabcolsep}{3pt}
  \resizebox{\columnwidth}{!}{%
  \begin{tabular}{@{}lrrrr@{}}
    \toprule
    Position & ECE $\downarrow$ & AUROC $\uparrow$ & WY Corr. $\uparrow$ & CY Dmg. $\downarrow$ \\
    \midrule
    \rowcolor{black!5}
    \multicolumn{5}{l}{\textit{Qwen3-4B}} \\
    No intervention & 21.9 & 66.5 & --- & --- \\
    Token 1 & 21.9 & 66.7 & 0.1 & 0.0 \\
    Token 2 & 21.9 & 69.6 & 0.2 & 0.0 \\
    Token 3 & 22.7 & 38.2 & 0.0 & 0.0 \\
    Token 4 & 15.1 & 88.1 & 35.3 & 3.8 \\
    Token 5 & 12.5 & 86.9 & \textbf{80.1} & 10.3 \\
    \textbf{Joint 1--5} & \textbf{9.2} & \textbf{91.1} & 63.3 & \textbf{5.1} \\
    \addlinespace
    \rowcolor{black!5}
    \multicolumn{5}{l}{\textit{Gemma3-4B}} \\
    No intervention & 26.3 & 76.8 & --- & --- \\
    Token 1 & 26.1 & 79.7 & 0.8 & 0.1 \\
    Token 2 & 26.3 & 79.5 & 0.5 & 0.0 \\
    Token 3 & 26.2 & 76.8 & 0.6 & 0.0 \\
    Token 4 & 26.1 & 80.5 & 0.9 & 0.0 \\
    Token 5 & 24.6 & 81.3 & 4.5 & 0.5 \\
    \textbf{Joint 1--5} & \textbf{23.6} & \textbf{82.1} & \textbf{9.5} & 0.7 \\
    \bottomrule
  \end{tabular}
  }
  \caption{Pathway intervention results on confidence-template tokens before the readout token. Single-position rows apply PCHI at one template token, while the joint row applies PCHI across tokens 1--5. All values are percentages, and no attention mask is used.}
  \label{tab:reach}
\end{table}

%% file: tables/table3.tex
\begin{table}[t]
  \centering
  \footnotesize
  \setlength{\tabcolsep}{3pt}
  \resizebox{\columnwidth}{!}{%
  \begin{tabular}{@{}lrrrr@{}}
    \toprule
    Visible context & ECE $\downarrow$ & AUROC $\uparrow$ & WY Corr. $\uparrow$ & CY Dmg. $\downarrow$ \\
    \midrule
    \rowcolor{black!5}
    \multicolumn{5}{l}{\textit{Token 1}} \\
    Full & 21.9 & 66.7 & 0.1 & 0.0 \\
    \textbf{Prompt} & \textbf{14.6} & \textbf{86.2} & \textbf{70.2} & 11.1 \\
    Reasoning & 15.4 & 78.4 & 28.7 & 5.4 \\
    Answer & 13.7 & 78.2 & 17.6 & 3.4 \\
    Reasoning+Answer & 22.5 & 78.6 & 0.4 & 0.0 \\
    \addlinespace
    \rowcolor{black!5}
    \multicolumn{5}{l}{\textit{Token 2}} \\
    Full & 21.9 & 69.6 & 0.2 & 0.0 \\
    \textbf{Prompt} & \textbf{13.2} & \textbf{88.3} & \textbf{80.6} & 11.0 \\
    Reasoning & 12.6 & 85.1 & 66.0 & 9.6 \\
    Answer & 14.6 & 83.5 & 33.0 & 7.7 \\
    Reasoning+Answer & 13.5 & 85.1 & 53.5 & 8.1 \\
    \addlinespace
    \rowcolor{black!5}
    \multicolumn{5}{l}{\textit{Token 3}} \\
    Full & 22.7 & 38.2 & 0.0 & 0.0 \\
    \textbf{Prompt} & \textbf{22.6} & \textbf{59.5} & \textbf{0.9} & 0.1 \\
    Reasoning & 22.8 & 32.7 & 0.0 & 0.0 \\
    Answer & 23.0 & 29.5 & 0.4 & 0.0 \\
    Reasoning+Answer & 22.7 & 38.3 & 0.0 & 0.0 \\
    \bottomrule
  \end{tabular}
  }
  \caption{Attention-mask analysis for Qwen3-4B on early confidence-template positions. Each row applies PCHI at the indicated token while restricting the visible context of post-probe template queries. All values are percentages.}
  \label{tab:mask}
\end{table}

%% file: tables/table4.tex
\begin{table}[t]
  \centering
  \footnotesize
  \setlength{\tabcolsep}{3pt}
  \resizebox{\columnwidth}{!}{%
  \begin{tabular}{@{}lrrrr@{}}
    \toprule
    Visible context & ECE $\downarrow$ & AUROC $\uparrow$ & WY Corr. $\uparrow$ & CY Dmg. $\downarrow$ \\
    \midrule
    Full & 23.6 & \textbf{82.1} & 9.5 & \textbf{0.7} \\
    Prompt & 21.4 & 80.0 & 18.6 & 2.4 \\
    Reasoning & 18.0 & 79.6 & 34.2 & 6.7 \\
    Answer & \textbf{17.3} & 79.5 & \textbf{40.4} & 6.1 \\
    Reasoning+Answer & 18.7 & 80.1 & 30.4 & 4.7 \\
    \bottomrule
  \end{tabular}
  }
  \caption{Gemma3-4B attention-mask analysis for joint early-token intervention. PCHI is applied across confidence-template tokens 1--5 while varying the visible context of post-probe template queries. All values are percentages.}
  \label{tab:gemma-mask}
\end{table}

%% file: tables/table5.tex
\begin{table}[t]
  \centering
  \footnotesize
  \setlength{\tabcolsep}{3pt}
  \resizebox{\columnwidth}{!}{%
  \begin{tabular}{@{}lrrrr@{}}
    \toprule
    Setting & ECE $\downarrow$ & AUROC $\uparrow$ & WY Corr. $\uparrow$ & CY Dmg. $\downarrow$ \\
    \midrule
    No intervention & 21.9 & 66.5 & --- & --- \\
    Mask only, Prompt & 22.6 & 62.0 & 1.0 & 0.1 \\
    PCHI, Token 1 + Prompt & 14.6 & 86.2 & 70.2 & 11.1 \\
    PCHI, Token 2 + Prompt & \textbf{13.2} & \textbf{88.3} & \textbf{80.6} & 11.0 \\
    PCHI, Token 3 + Prompt & 22.6 & 59.5 & 0.9 & 0.1 \\
    \bottomrule
  \end{tabular}
  }
  \caption{Mask-only ablation on Qwen3-4B. The prompt-only mask is compared against PCHI under the same visible-context restriction. Masking alone does not explain the large correction gains at tokens 1 and 2, while token 3 remains close to the mask-only behavior. All values are percentages.}
  \label{tab:mask-only}
\end{table}

%% file: latex/conclusion.tex
\section{Conclusion}
\label{sec:conclusion}
This paper treats verbalized confidence not only as an output to be
recalibrated, but as an internal computation that can be selectively controlled.
We introduced probe-conditioned head intervention: a frozen probe estimates, at
a confidence-template position, whether a confident response is likely wrong,
and a learned head-level intervention rescales downstream attention-head outputs
only in proportion to that evidence. On structured mathematical self-evaluation,
this lets the model suppress many unwarranted \texttt{yes} confidence readouts
while preserving most warranted ones.

The results suggest a more fine-grained view of overconfidence correction. At
the readout token, wrong-yes confidence can be directly suppressed,
most strongly on Qwen3-4B. At upstream confidence-template tokens, the effect depends on
position, model, and visible context, showing that confidence-relevant evidence
is not uniformly usable along the answer-to-readout pathway. Together with the
mask-only ablation, these findings indicate that the gains are not simply a
byproduct of restricting attention or globally lowering confidence, but arise
from evidence-conditioned changes to downstream head computation.

More broadly, PCHI shows that calibration need not be treated only as a post-hoc
mapping from scores to probabilities. When confidence-relevant evidence is
available inside the model, selective internal interventions can reduce
unwarranted confidence while preserving much of the confidence assigned to
correct answers. Extending this idea beyond structured binary confidence fields
is an important next step toward models whose expressed uncertainty is both
useful and trustworthy.

\section*{Limitations}

Our experiments are limited to mathematical question answering with a structured JSON output format and a binary \texttt{yes}/\texttt{no} confidence field. This setting makes the confidence template explicit and allows the template tokens to be fixed during inference. The method has not yet been tested on free-form confidence expressions, multi-level confidence scales, or open-ended tasks where the boundary between answer generation and confidence assessment is less controlled.

The intervention also depends on model- and data-specific probe training. Probe layers, intervention layers, template positions, and attention masks are selected from diagnostics on the training distribution, and different models show different effective positions and correction-damage trade-offs. Extending the approach to broader model families and task distributions will require testing how stable these choices are and whether they can be selected automatically.

Finally, PCHI provides evidence of selective control, but it is not a complete mechanistic explanation of verbalized confidence. Probe separability shows that wrong-confident evidence is decodable at confidence-template positions, and the intervention results show that downstream head outputs can alter the final confidence readout. The learned scalar coefficients do not by themselves explain how individual heads encode or combine correctness-related information, leaving circuit-level analysis as an important direction for future work.

%% file: latex/appendix.tex
\section{Implementation Details}
\label{sec:appendix}
\subsection{Generation Prompt}
\label{sec:appendix-prompt-template}
\begin{promptbox}
You are solving a math problem. Respond ONLY with valid JSON in this exact format:

{"reasoning": "brief step-by-step explanation", "answer": "final answer LaTeX only", "is_confident": "yes or no"}

### Rules:
- "reasoning": Your brief thought process
- "answer": ONLY the final result itself. Do NOT wrap it in any extra command or delimiter.
- Use LaTeX when the answer is a fraction, radical, set, interval, equation, or expression.
- Because this is JSON, escape backslashes in LaTeX. Example: "\\frac{3}{8}".
- "is_confident": Answer "yes" if confident, "no" if uncertain.

### Invalid (DO NOT do this):
- Do NOT include "The answer is" in answer field
- Do NOT use any wrapper around the answer
- Do NOT put reasoning or explanation in answer field
- Do NOT write anything outside the JSON
- Do NOT make reasoning over 1000 tokens.

### Important:
- If you cannot solve the problem, do not pile up meaningless reasoning; honestly acknowledge that you cannot give a certain answer.
- When filling the "is_confident" field, honestly self-assess whether your reasoning is reliable and whether the answer is trustworthy.

### Question:
{question_text}

### JSON Response:
\end{promptbox}
\subsection{Probe Diagnostics and Layer Selection}
\label{sec:appendix-probe-diagnostics}

\begin{figure}[t]
\centering
\includegraphics[width=\columnwidth]{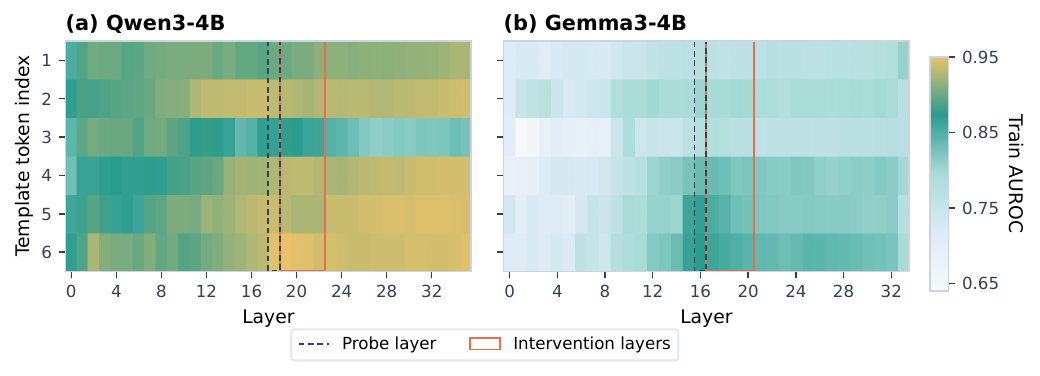}
\caption{Training-set AUROC of WY-vs-CY confidence-template probes across template positions and transformer layers. Dashed outlines mark the selected probe layers, and solid outlines mark the downstream intervention layers.}
\label{fig:probe-auroc-heatmap}
\end{figure}

Figure~\ref{fig:probe-auroc-heatmap} reports the training-set AUROC of the
WY-vs-CY probes across confidence-template positions and transformer layers. We
use training-set diagnostics for layer selection so that the validation split
remains reserved for final intervention evaluation. The heatmaps make the layer
selection criterion explicit: the intervention can only act on
wrong-vs-correct-confident evidence that is already linearly decodable at the
probe layer, so we look for a layer where WY/CY separability is high while
several downstream layers remain available for intervention in the same forward
pass.

Two features of the heatmaps drive the choice. First, separability is not
present from the start: in both models the early layers are close to chance, and
the AUROC rises into the middle of the network before saturating. This is why
the probe layer is placed in the middle rather than early -- earlier layers do
not yet carry a reliably decodable wrong-confident signal, and intervening
before that signal exists would have nothing to condition on. Second, because we
restrict the intervention to layers strictly after the probe layer (so that the
hidden states the probe consumes stay on their training distribution), the probe
layer must be early enough to leave a usable band of downstream layers. The
selected layers satisfy both constraints: for Qwen3-4B, layer 18 combines strong
separability across the template with four clean downstream layers, so we probe
at layer 18 and intervene at layers 19--22; for Gemma3-4B, the analogous point is
layer 16, with intervention at layers 17--20.

The heatmaps also explain two results reported in the main text. The Qwen3-4B
token-3 row is visibly weaker than its neighbors (tokens 2 and 4) across most
layers, indicating that this template coordinate carries a weaker
wrong-confident signal; this lower separability is consistent with token 3 being
an ineffective intervention point in Table~\ref{tab:reach}, even under attention
masking. The token corresponds to a low-information position in the
confidence-template span ("\_conf"), which
plausibly accounts for its weak separability relative to adjacent positions.
More broadly, Gemma3-4B is systematically less separable than Qwen3-4B: its
AUROC values are lower across nearly all positions and layers. This weaker and
less peaked separability is consistent with the smaller and more
context-dependent intervention effect we observe for Gemma throughout
Section~\ref{sec:reach}: when the wrong-confident signal available to the probe
is weaker, the conditional head intervention has correspondingly less evidence to
act on.

